\pdfoutput=1
\documentclass[11pt]{article}

\usepackage[final]{coling}

\usepackage{times}
\usepackage{latexsym}

\usepackage[T1]{fontenc}
\usepackage[utf8]{inputenc}

\usepackage{microtype}

\usepackage{inconsolata}

\usepackage{graphicx}

\usepackage{soul}
\usepackage{url}
\usepackage[utf8]{inputenc}
\usepackage{graphicx}
\usepackage{amsthm}
\usepackage{booktabs}
\usepackage{algorithm}
\usepackage{algorithmic}
\usepackage[switch]{lineno}
\usepackage{rotating}
\usepackage{lscape}
\usepackage{booktabs}
\usepackage{amsfonts,amsmath,amssymb,bm}
\usepackage{setspace}
\usepackage{pdflscape}
\usepackage{pdfpages}
\usepackage{mathabx}
\usepackage{subcaption}
\usepackage{mathtools}
\usepackage{microtype}
\usepackage{xcolor}
\usepackage{tcolorbox}
\usepackage{enumitem}
\usepackage{multirow}
\usepackage{caption}
\usepackage{subcaption}
\usepackage{graphicx}
\usepackage{adjustbox}

\title{Automatic Extraction of Metaphoric Analogies from Literary Texts: \\
Task Formulation, Dataset Construction, and Evaluation}

\author{Joanne Boisson$^1$,  \textbf{Zara Siddique}$^1$, Hsuvas Borkakoty$^1$, \\
\textbf{Dimosthenis Antypas}$^1$, \textbf{Luis Espinosa Anke}$^{1,2}$ \and \textbf{Jose Camacho-Collados}$^1$ \\
  $^1$Cardiff NLP, School of Computer Science and Informatics, Cardiff University, U.K.\\
  $^2$Amplyfi, Cardiff, U.K. \\
  \texttt{\{boissonjc,siddiquezs2,borkakotyh,camachocolladosj\}}
  \\ \texttt{@cardiff.ac.uk} \\
  \\}

\begin{document}
\maketitle
\begin{abstract}
  Extracting metaphors and analogies from free text requires high-level reasoning abilities such as abstraction and language understanding. Our study focuses on the extraction of the concepts that form metaphoric analogies in literary texts. To this end, we construct a novel dataset in this domain with the help of domain experts. We compare the out-of-the-box ability of recent large language models (LLMs) to structure metaphoric mappings from fragments of texts containing proportional analogies. The models are further evaluated on the generation of implicit elements of the analogy, which are indirectly suggested in the texts and inferred by human readers. The competitive results obtained by LLMs in our experiments are encouraging and open up new avenues such as automatically extracting analogies and metaphors from text instead of investing resources in domain experts to manually label data.
\end{abstract}

\section{Introduction}

According to \citet{hofstadter2013surfaces}, analogy is the fuel and fire of thinking because humans dynamically build concepts by analogy. Although it is a core mechanism of the mind, they have proven to be difficult to extract automatically from free text, 
because they can involve some implicit concepts and can link dissimilar concepts. In some cases, in particular when analogies pair very different concepts, they can also be metaphoric \cite{gentner-career}.
For example, a \textit{head} and an \textit{apple} can be mapped in the sentence \textit{My head is an apple without a core} \cite{Sternberg_Tourangeau_Nigro_1993}, and form an analogy that is also a metaphor. The recent progress of transformer-based large language models (LLMs) open a path towards a finer-grain semantic handling of metaphoric analogies in Natural Language Processing. 

Reasoning by analogy at a human like level is an ability pointed out by \citet{DBLP:journals/corr/abs-2102-10717} as lacking in LLMs. These reasoning capabilities allow humans to learn from few examples, and draw from relevant past experiences to tackle new problems. Analogical thinking is a process of generalization and abstraction. Large pretrained language models have some analogical abilities, they can be prompted successfully for analogical reasoning \cite{yasunaga2024largelanguagemodelsanalogical}, and  perform zero-shot analogical reasoning in visual task after converting them into language \cite{hu-etal-2023-context}. Do they also recognize complex metaphoric analogies in texts?

~\citet{webb-emergent} solve a broad range of different analogies, from geometric patterns to short pieces of text, by prompting the first GPT-3 instructed model ~\cite{GPT3,ouyang2022training}.
~\citet{tong2024metaphor} release new resource and baselines for metaphor understanding tasks. Beyond the field of NLP, in linguistics, psychology or neuroscience, research on metaphors in the last five decades have been growing in importance \cite{Peng2023-qp}.  
With the new NLP state of the art, more attention is now given to analogy processing in Artificial Intelligence\footnote{https://analogy-angle.github.io/}, with the idea of bridging the gap towards other disciplines. For example, researchers have attempted comparisons of the information or features on which humans and models rely to solve analogies ~\cite{stevenson2023large}. To our knowledge, there is currently no evaluation of the ability of LLMs to extract mappings involving more than one single pair of concepts from text.

\begin{table*}[h!]
\footnotesize
\begin{tabular}{l@{\hspace{5pt}}c@{\hspace{5pt}}c@{\hspace{5pt}}c@{\hspace{5pt}}l@{\hspace{5pt}}l}
\toprule
\textbf{Text}&T1&T2&S1&S2&i\\
\midrule

\begin{tabular}{l@{\hspace{1pt}}} \textbf{Books} \underline{are like} \textbf{imprisoned souls} till someone takes them \\ down from a \textbf{shelf} and frees them. (S. Butler)\end{tabular} &
\begin{tabular}{l@{\hspace{1pt}}}books\end{tabular}&
shelf&
\begin{tabular}{l@{\hspace{1pt}}}(imprisoned)\\souls\end{tabular}&
\(<\)prison\(>\)&
1\\\midrule

\begin{tabular}{l@{\hspace{1pt}}} \textbf{Money} \underline{is} the \textbf{mother's milk} of \textbf{politics}. (J. Unruh)\end{tabular}&
\begin{tabular}{l@{\hspace{1pt}}}money\end{tabular}&
politics&
\begin{tabular}{l@{\hspace{1pt}}}(mother's)\\milk\end{tabular}&
\(<\)baby\(>\)&
1\\\midrule

\begin{tabular}{l@{\hspace{1pt}}} An election is coming. Universal peace is declared,\\ and the \textbf{foxes} have a sincere interest \\in prolonging the lives of the \textbf{poultry}. (G. Eliot)\end{tabular} &
\begin{tabular}{l}\(<\)candidates\(>\)\end{tabular}&
\(<\)voters\(>\)&
\begin{tabular}{l@{\hspace{1pt}}}foxes\end{tabular}&
poultry
&2\\

\bottomrule
\end{tabular}
\caption{Example of metaphors and their extracted analogical frames: the pair of target domain concepts (T1 \& T2), the pairs of source domain concepts (S1 \& S2) and the number of implicit terms of the analogy (i). Implicit terms appear between brackets (<>). Modifiers of the main concepts of the analogy appear between parenthesis.}
\label{tab:examples}
\end{table*}

Analogies are parallels or mappings across concepts. Proportional analogies, which are the focus of this study, are an association between two pairs of concepts such as \textit{Answer is to riddle what key is to lock}.

We tackle the question of metaphorical mapping identification at the lexical level, by extracting the core words involved in an analogy from pieces of text and prompting the models to output the eventual implicit concepts involved.
For example, the poem verse \textit{Memory, a jar of flies. Spin off the lid.} (Seibles, 1955) maps \textit{memory} to a \textit{jar}, and \textit{flies} to an implicit concept that could very well be \textit{<recollections>}. 

Given a short text containing a metaphor, the task we introduce is to extract a pair of Noun Phrases (NPs) belonging to the source domain of the metaphor, i.e. expressions that are used metaphorically (\textit{jar} and \textit{flies} in our example) and another pair of NPs belonging to the target domain, i.e. the topic being discussed (\textit{memory} and \textit{<recollections>}, such as \textit{flies} are to the \textit{jar} what \textit{recollections} are to \textit{memory}). We view this task as a step towards the extraction and structuring of more complex metaphoric mappings from free texts. Such research direction opens the path toward the creation of large scale knowledge base of metaphoric mappings that could be benefit disciplines such as neuroscience, cognitive science, sociology, anthropology or literature.

Elucidating freely expressed complex analogies also has the potential to improve NLP downstream tasks such as Question Answering, Natural Language Inference or Machine Translation \cite{li2024findingchallengingmetaphorsconfuse}. This task can provide greater value to NLP systems than merely identifying words used metaphorically in texts, because more information is extracted when both the source and target domains are modeled. For example, indirect biases \cite{DBLP:journals/corr/BolukbasiCZSK16a}, which extend beyond gender stereotypes and can sometimes manifest as subtle hate speech, can be more effectively identified and addressed through analogy extraction.

Our main contributions are the following. We release a dataset of 204 instances composed of sourced metaphors\footnote{Most have at least an author and the Metaphor of Mind instances have a complete bibliographic reference.}, labelled for source and target concept pairs. We evaluate and compare recent transformer-based language models on a new task: identifying the elements of 4-term metaphoric analogies in literary English texts. We then manually evaluate the ability of the models to generate relevant concepts when one or more of the four elements of the analogy are implicit.

\section{Task motivation and definition}

Our study takes root in the observation that recent large pre-trained language models often explain complex literary metaphors and similes well, as in the example below, where the correct intentionality behind the statement is also stated

\begin{quote}
There are fascists pretending to be humanitarians like cannibals on a health kick eating only vegetarians. (Roger McGough)
\end{quote}

OpenAI GPT-3.5 generates this explanation:

\begin{quote}
[...] It highlights the hypocrisy and deceitfulness of such behavior, likening it to cannibals who, while ostensibly trying to improve themselves, are still engaged in morally reprehensible actions.
\end{quote}

Large collections of literary metaphors have been compiled, such as \citet{metaphorsofmind:2015}, but they do not usually contain annotations of the exact words that are mapped to one another in the text (e.g. \textit{fascists} -> \textit{cannibals on a health kick}), or are limited to tagging a single pair of concepts per instance \cite{mohler-etal-2016-introducing}. A strong motivation behind this work is to take a first step toward leveraging existing resources and enriching them with mapped concepts and relations that could benefit further metaphor and analogy studies.

We design a task that involves (1) the extraction of the explicit elements that form an analogy via two pairs of concepts, (2) the identification of the source and target domain of the metaphor, and (3) the generation of relevant concepts to fill the missing frames of a mapping when they are implicit in the original text. We focus on short famous or literary texts\footnote{Meaning that the metaphors have probably been seen by the LLMs at pretraining time, but are not always conventional.}, ranging from a sentence to a paragraph, that all contain a metaphoric analogy. These are proportional analogies involving two pairs of concepts, that have the conceptual form \(T1:T2 :: S1:S2\), to be interpreted as \textit{T1 is to T2 what S1 is to S2}, where T1 and T2 are concepts belonging to the target domain of the metaphor, i.e. the topic being discussed in the text, and S1 and S2 belong to the source domain of the metaphor, i.e. the concepts used metaphorically in the text. Table \ref{tab:examples} shows input examples for the task on the left, and the desired extracted output (T1, T2, S1 and S2).

\section{Related work}

In this section, we first discuss theories about the relations between metaphors and analogies (\ref{sec:metaphors-and-analogies}), and then review previous work related to metaphoric mapping in NLP,  before narrowing the scope to approaches to source and target metaphoric mapping identification in natural text (\ref{sec:review-mapping}).

\subsection{Analogies and metaphors}
\label{sec:metaphors-and-analogies}
Metaphors are the subject of a multidisciplinary field of research of their own, with corpus based analysis in various disciplines in humanities, as well as research in cognitive sciences, psychology, linguistics and philosophy of language. 
 The relation between metaphors and analogies has been much debated. Researchers who refer to shared features and structural analogies as the basis of metaphors disagreed with conceptual mapping theorists who argued that similarity is not the basis for metaphors \cite{grady-1999}.
 \citet{gentner-like} and \citet{gentner-career} introduce a framework that intends to unify both views. 
 Following their theory, we consider metaphors as a subset of analogies, and include similes in this subset. The question of how metaphors are processed by humans or machines is beyond the scope of our work. According to the taxonomy of \citet{wij-neurosymbo-2023}, the metaphors included in our dataset are \textit{semantic and pragmatic analogies}, i.e. the two most complex types of analogies, which require good syntactic and semantic representations, and sometimes pragmatic knowledge.

Metaphors are a mapping of concepts from two domains. The concepts and relations from a target domain are described in terms of concepts and relations from the source domain. Source and target domains are often semantically distant, many metaphors are far analogies, which makes them harder to process for LLMs \cite{ushio-etal-2021-bert}. They often involve an abstract target domain and a concrete source domain. Therefore, unlike some other analogies, the mapping is not reversible \cite{ortony_1993}, i.e., metaphors have directionality. For example \textit{The acrobat is a hippopotamus} suggests a clumsy acrobat and \textit{The hippopotamus is an acrobat} suggests a graceful hippopotamus.

\subsection{Metaphoric mappings}
\label{sec:review-mapping}

\citet{LakoffJohnson80}'s Conceptual Metaphor Theory (CMT) demonstrates how everyday language is filled with metaphors that we may not always notice. Conceptual mappings shape the way we think and are shown through the way we speak. NLP researchers have designed models based on this theory, either to discover large conceptual mappings between domains, or on the contrary, used a set of mappings as latent variables for figurative language analysis or generation.

\citet{mason-2004-cormet}'s CorMet System discovers metaphorical mappings between concepts using pairs of domain corpora. It does not try to identify mappings at the sentence level, it searches for the characteristics of a metaphoric mapping from an aggregation of verb-object features. A few years later, \citet{shutova-etal-2010-metaphor} introduced pre-defined mappings to cluster instances of verb-object pairs.

In the area of metaphor generation, \citet{veale-li-2012-specifying} develop a system of metaphor interpretation and generation, \textsc{Metaphor Magnet}, that relies on the harvest of stereotype from the Google n-grams. The stereotypical attributes associated to concepts are leveraged for the suggestion of relevant metaphors and their interpretation (e.g. given \textit{Google is –Microsoft}, the system outputs \textit{giant} with properties like \textit{lumbering} and \textit{sprawling}).~\citet{stowe-etal-2021-metaphor} rely on conceptual mappings to guide the generation of metaphors in context. Mappings are extracted with a methodology using FrameNet \cite{baker-etal-1998-berkeley-framenet}.

More work has been done on metaphoric mappings such as
~\citet{sultan-shahaf-2022-life} who map pairs of short texts involving metaphors, \citet{10.1162/tacl_a_00688} who create  a benchmark for Analogical Reasoning on Narratives (ARN) distinguishing between near and far analogies, and ~\citet{czinczoll-etal-2022-scientific} who released a corpus of metaphoric analogies in the form of quadruples. ~\citet{survey2022} publish two surveys on metaphor processing including a review on mapping related research.

\paragraph{Everyday metaphors and literary metaphors} Our choice to focus on literary metaphors may appear restrictive, unlike recent work that pivots towards Lakoff's theory focusing on everyday language and is anchored in research on metaphors beyond the scope of stylistics. We are motivated by the search for collections of proportional analogies in natural language, with the constraint of avoiding the harvest of candidate sentences from large corpora using pattern-based queries, that would create a strong bias in the surface form taken by the analogies. \citet{metaphorsofmind:2015} and \citet{grothe} are two collections of metaphors that contain relevant candidate literary texts. We live by them as they are a cultural inheritance, especially the most famous ones such as William Shakespeare's \textit{world is a stage} quote. They surround us and reappear under various surface forms in everyday language.

\paragraph{Detection of source-target concept pairs in corpora} There is a line of work aiming at deriving domain mappings from single sentences. Some research focused on specific types of metaphors in corpora, such as synesthesia, with ~\citet{jiang-etal-2022-chinese} in the Mandarin Chinese Sinica Corpus for the detection  and analysis of synesthesia.

For general mapping inference, ~\citet{Ge_Mao_Cambria_2022} design a model deriving a source and target concept pair by abstracting over a metaphoric pair of syntactically related words. Similarly, ~\citet{wachowiak-gromann-2023-gpt} extract mappings from metaphoric sentences using GPT-3 and existing annotated datasets for source and target domains such as the LCC ~\cite{mohler-etal-2016-introducing}.

We do not abstract over the concepts that appear in our dataset sentences with clustering or generalization through hypernym identification. We instead explore the out of the box potential of recent large language models to directly reconstruct analogical structures using the NPs that appear in the texts. 

\section{Dataset construction}
\label{sec:dataset-construction}
The dataset construction consists of manually selecting short texts containing 4-term metaphoric analogies from existing collections, the tagging of the NPs corresponding to the frames of an analogy, and the suggestion of a relevant concept by an annotator to identify unstated implicit terms.

\subsection{Example selection}
\label{sec:example-selection}
\paragraph{Data sources}The examples are collected from the Metaphor of Mind  project repository\footnote{\url{https://metaphors.iath.virginia.edu/metaphors}} \cite{metaphorsofmind:2015}, two books that compile famous metaphors ~\cite{grothe,book-of-similes} and a few sourced metaphors found online\footnote{For example, \url{https://prowritingaid.com/metaphor-examples}}.
These are extracts of novels, poems, songs or famous speeches with a known author.

\paragraph{Selection of proportional analogy examples} 
Metaphors that do not naturally fit the proportional analogies between pairs of concepts and one relation, i.e., \(T_1: T_2 :: S_1 : S_2\), are discarded. For example, the metaphors \textit{If you torture the data long enough, it will confess to anything.} (Ronald Coase) implies three source concepts (the torturer, the prisoner and the confession), three target concept (you, the data and false conclusions)  and two relations (torture and confess) that have different arguments. Keeping such metaphors in our dataset would require an adaptive structure mapping method allowing the mapping of triples, or would inject more ambiguity in the possible correct answers within our current mapping structure. Smaller one-to-one mappings such as \textit{Silence is a black cavern} \cite{katz-norms}, could be expanded into more complex ones to fit our framework (e.g. a \textit{friendly voice} is to \textit{silence} what a \textit{candle} is to a \textit{black cavern}), but there are many ways to enrich them with possible implicit terms, also adding ambiguity to the task.

\paragraph{Removal of nested metaphors} During the data collection, we avoid multiple levels of nested metaphors or juxtaposed metaphors to limit the risk of confusion given the prompts in our experiments. For example, in the following example, \textit{taxation} is compared to \textit{plucking the goose} and to an \textit{art}: \textit{The art of taxation consists in so plucking the goose as to obtain the largest amount of feathers with the least amount of hissing.} (Jean-Baptiste Colbert). Two source domains are thus mapped to \textit{taxation}, which would introduce confusion in our task, so we discard this example and leave this case to future work.

\paragraph{Implicit analogies}

We retain instances that involve concepts S1, S2, T1 or T2 that are not explicitly mentioned in the text. For example the metaphor, \textit{My head is an apple without a core} is missing a term \cite{Sternberg_Tourangeau_Nigro_1993}. Comprehending the metaphor requires understanding the relation between \textit{head} and \textit{apple}, and also inferring the relation between \textit{apple} and \textit{core}, in order to complete the analogy with the missing implicit term \textit{brains} (or a similar term analogous to \textit{core}).

In Table \ref{tab:examples}, the first example is completely explicit, i.e. the four concepts forming the analogy are given in the text. In the second example, one term of the source domain, \textit{baby}, is implicit. In the third example, the two target concepts forming the mapping are implicit and other nouns present in the sentence might confuse the model.

\begin{table}[t!]
\footnotesize
\begin{tabular}{ll@{\hspace{4pt}}l@{\hspace{7pt}}l@{\hspace{7pt}}l@{\hspace{7pt}}l@{\hspace{14pt}}l@{\hspace{7pt}}l@{\hspace{7pt}}l@{\hspace{7pt}}l@{\hspace{7pt}}l}
\toprule

\textbf{An.}&\multicolumn{5}{c}{\textbf{Exact match}}&\multicolumn{5}{c}{\textbf{Head noun match}}\\

&\textbf{T2}&\textbf{S1}&\textbf{S2}&\textbf{All}&\textbf{Q}&\textbf{T2}&\textbf{S1}&\textbf{S2}&\textbf{All}&\textbf{Q}\\
\midrule

\textbf{A1}& .57&	.71&	.46& .58&.28& .57 &.71 & .47& .58&.29\\
\textbf{A2}& .51&	.76&	.58& .62&.31&.51 &.76 &.64&.64 &.35\\
\textbf{A3}& .68&	.84&	.50& .67 &.38& .68&.84 &.51& .68&.40\\
\textbf{A4}& .70&	.87&	.62& .73&.51 & .70& .87&.67 &.75&.54\\
\textbf{A5}& .72&	.87&	.62&.74&.53& .72 & .87&.65& .74&.54\\
\midrule
\textbf{Av.}&.64 &.81 & .56&.67 &.40 &.64 &.81 &.59 &.68&.42\\
\bottomrule
\end{tabular}
\caption{Averaged pairwise inter-annotator agreement scores with Cohen's Kappa. Agreement by frame, summed for all the frames (All) and by quadruple (Q). Exact match and lemmatized head-noun based match.}

\label{tab:inter-annotation-agreement}
\end{table}

\subsection{Annotation}

The selection and annotation of 204 instances is carried out by one author who is an expert in metaphors and is double-checked by a second author. The reason for this choice lies in the limited budget as well as the complexity of the task. We also perform an inter-annotator agreement test that highlights the variances among different annotators. A summary of the dataset can be found in Table \ref{tab:dataset-description}.

\paragraph{Inter-annotator agreement (IAA) metrics} 20 instances are labeled by five annotators with a background in metaphors studies, linguistics or literature. After studying a few examples, given a short text and a target concept T1, they are asked to find the three other concepts forming an analogy with T1. The concepts must be syntagms with nominal heads.\footnote{We found that this instruction helped the annotators to separate the relations from the arguments of the analogy.} The details of the inter-annotator agreement test and results by frame are provided in Table \ref{tab:inter-annotation-agreement}. The implicit terms of the analogy are replaced with a unique token \(<\)\textit{implicit}\(>\)  before computing the agreements.\footnote{Table \ref{tab:appendix-iaa-without-implicit} in the Appendix shows an alternative measurement where the frame is excluded from the calculation if a majority of annotators marked it as an implicit frame.} 

\paragraph{IAA results} On average 4 of the 5 annotators give identical answers for the frame T2, 4.5 for the first source concept S1 and 3.8 for S2 of the analogy. Annotators tend to agree more on the term S1, which happened to always be explicit and often designated with comparison markers in our small IAA test data. In addition to the agreement by frame, Table \ref{tab:inter-annotation-agreement} also shows the agreement for complete quadruples (in the column Q), which is quite weak, revealing the difficulty of the task. Annotators A4 and A5 reach a strong pairwise agreement for all the frames. This suggests that the task is not highly ambiguous, but that disagreement is caused by the difficulty in finding a satisfying solution for several sentences, as reported by three annotators.

\paragraph{Expert annotator} The author who selects the analogies and labels for the entire dataset, annotator A5 in Table \ref{tab:inter-annotation-agreement}, has on average the highest pairwise Cohen's kappa score with the other annotators. Annotator A5 answers to the analogy extraction task are in the majority vote \(<\)95\% of the time.

\paragraph{Dataset}
Table \ref{tab:dataset-description} shows a summary of the resulting dataset.\footnote{As well as Table \ref{tab:implicit-terms} in the Appendix.} The majority of the instances have 0 or 1 implicit terms, evenly distributed. A small subset of 10 instances contain two implicit terms, one such example is shown in the last row of Table \ref{tab:examples}. The majority of implicit terms are concentrated in the S2 frame of the source domain. The four arguments of each analogy have different functions in context: we label T1 the main target concept of analogy (often also the subject of the main sentence). It is often explicitly mapped to S1 that also has a low rate of implicit terms. In contrast, T2 and S2 may be only implicitly suggested.
The selected texts have variable lengths. The average token length of the instances selected from the ~\citet{metaphorsofmind:2015} corpus of Metaphors of Mind is 29.9, longer than the texts selected from the ~\citet{grothe} collection with 17.7 tokens on average\footnote{The distribution of the number of nouns per data source can be seen in the Appendix Section \ref{appendix-histograms}.}.

\begin{table}[t!]
\footnotesize
\begin{tabular}{l@{\hspace{5pt}}c@{\hspace{5pt}}c@{\hspace{5pt}}c@{\hspace{6pt}}c@{\hspace{5pt}}c@{\hspace{5pt}}c}
\toprule
\textbf{Source}&
\begin{tabular}{l@{\hspace{5pt}}}\textbf{n}\\\textbf{ins.}\end{tabular}&
\begin{tabular}{c@{\hspace{6pt}}}\textbf{i=0}\\\textbf{(\(\%\))}\end{tabular}&
\begin{tabular}{l@{\hspace{6pt}}}\textbf{i=1}\\\textbf{(\(\%\))}\end{tabular}&
\begin{tabular}{l@{\hspace{6pt}}}\textbf{i=2}\\\textbf{(\(\%\))}\end{tabular}&
\begin{tabular}{l@{\hspace{5pt}}}\textbf{av.}\\\textbf{tok.}\end{tabular}&
\begin{tabular}{c@{\hspace{5pt}}}\textbf{av.}\\\textbf{N}\end{tabular}\\

\midrule
Grothe&124&60 (48)&61 (49)&3 (2)&17.7&4.7\\
Pasan.&60&27 (45)&25 (42)&8 (12)&29.9&6.8\\
Other&20&12 (60)&8 (40)&0 (0)&17.8&4.3\\
\midrule
\textbf{Total}&204&99 (48)&94 (46)&11 (5)&21.3&5.3\\
\bottomrule
\end{tabular}
\caption{Dataset summary for metaphors from our different sources. The columns show the number of instances (n ins.), the number of examples with \(i\) implicit terms (e.g. i=0 indicates that the four-term analogy has 0 implicit terms -- in other words, it only explicit terms mentioned in the metaphors), the average number (av.) of tokens (tok.) and nouns (N) per instance.}
\label{tab:dataset-description}
\end{table}

\section{Experimental Setting}

For the following experiment, the extraction of concepts from the text and generation of implicit terms is performed in a single step. 

\subsection{Analogy extraction}
\label{sec:analogy-extraction}

\paragraph{Task} Given a text and one of the four concepts with its frame (e.g. \textit{S1:apple}), we wish to extract the three other elements forming a metaphoric mapping. We run the experiments for each possible input concept of the quadruples and for all instances of the test set.
The prompt used for our experiments can be found in Appendix \ref{appendix-prompts}. The prompt aims to describe the task similarly to how it was explained to the annotators, and additionally includes seven examples to guide the model.

\paragraph{Models} We evaluate the following instructed versions of large decoder-only transformer-based language models:
Meta-Llama-3-70B \cite{touvron2023llama}\footnote{\url{https://huggingface.co/meta-llama/Meta-Llama-3-70B}}, GPT-3.5, GPT-4 \cite{bubeck2023sparks}. Sparse Mixture of Experts Mixtral models \cite{jiang2024mixtral} 8*7B and 8*22B parameters are also included in the evaluation\footnote{\url{https://huggingface.co/mistralai}}. The GPT models are queried through the API and the open source models are downloaded from Huggingface \footnote{\url{https://huggingface.co/}}. We run Llama-3 with an additional system prompt to better control the output format, the details are shown in Appendix \ref{sec:appendix-llama3-prompt}.\footnote{The prompt for Mixtral models is also adapted with the insertion of recommended special tokens for instruction \([INST]\), \([/INST]\), \(<\)s\(>\) and \(<\)/s\(>\) for instruction/answer pairs delimitation.}
For the large versions of Mixtral and Llama-3, we use quantization with the the bitandbytes library\footnote{\url{https://pypi.org/project/bitsandbytes/}} to reduce the computational load. 
All experiments are run with the models' default parameters including temperature.

\paragraph{Example sampling for in-context learning} We prompt the models in a few-shot setting, providing several batches of seven examples randomly sampled from our dataset. For each run, the seven examples used for in-context learning input are removed from the test set reducing it to 197 instances. Each short text is provided once along with each frame value (T1, T2, S1 or S2), including the frames that are have implicit values in the text, and for which values were given during the annotation process.
We run the experiment on GPT-3.5-turbo with 10 different batches of distinct examples and look at the standard deviation of the accuracy. GPT-4, Llama-3 and Mixtral models are then compared with GPT-3.5 on the same 3 batches of examples. We average the performance across the three batches for each model.

\subsection{Evaluation methods}

\paragraph{Evaluation metrics for NP extraction} The length of extracted strings may vary when obtaining correct analogy frames from models. To address this difficulty, we test four simple term-matching functions between the manually tagged element and the expressions extracted by the models. After lower-casing the text, we compute the exact match, overlap (if one string is included in the other), lemmatized expressions and lemmatized head noun matching metrics. The frame and value given in the input queries are discarded from the accuracy computation. 

\paragraph{Human judgment of the generated implicit terms}
We manually evaluate generated implicit terms for the two best performing models in our experiments, GPT-4 and Mixtral-8*22-instruct. We consider the set of sentences that have one implicit term among the four concepts of the analogy, and for which both models have successfully extracted the two explicit concepts of the quadruple according to the head-noun based accuracy metric.\footnote{The fourth term is provided in the prompt for the extraction task.} A set of twenty sentences is first labeled by two annotators for inter-annotator agreement validation. Annotators are asked to rate the generated expression from 0 to 2: 0 being incorrect, 1 making sense but being imperfect and 2 being very good, as exemplified in Table \ref{tab:examples-argument-generation}. We find a \textit{substantial} inter-annotator agreement according to \citet{landis-1977}'s scale
(Spearman $\rho=0.7$, $pvalue=1.88e-19$). Fifteen additional sentences are then judged by each annotator, bringing the total number of evaluated sentences to 50.

\paragraph{Implicit terms evaluation metric} A test instance with correctly extracted explicit terms by one given model might appear up to nine times within this evaluation: inference on a sentence is run for 3 batches of different examples and for 3 possible anchor concepts in the input. On average, the sentences used for evaluation appear 6.85 times for Mixtral 8*22 and 7.1 times for GPT-4 in the data. Table \ref{tab:examples-argument-generation} shows examples of several generated implicit concepts for one sentence.

When judging the results for 20 randomly selected sentences for two models, the initial set of instances contains 271 generated terms with duplicates. After deduplication, 122 unique generated terms and sentence pairs remain. For the 50 sentences in total, 725 generated terms and 291 distinct items are rated. Our metric is the results of averaging multiple ratings for each sentence for a model. A sentence is attributed with one final score per model.

\begin{table}[t!]
\footnotesize
\centering
\begin{tabular}{l@{\hspace{3pt}}l@{\hspace{5pt}}l@{\hspace{5pt}}l@{\hspace{5pt}}l@{\hspace{14pt}}l@{\hspace{5pt}}l@{\hspace{5pt}}l@{\hspace{5pt}}l}
\toprule
&\multicolumn{4}{c}{\textbf{Frame-blind}}&\multicolumn{4}{c}{\textbf{Frame-wise}}\\
\textbf{Model}&\textbf{raw}&\textbf{ovlp}&\textbf{lem}&\textbf{hd}&\textbf{raw}&\textbf{ovlp}&\textbf{lem}&\textbf{hd}\\
\midrule
Llama-3.inst &.69&.83&.72&.81&.61&.72&.63&.71\\
Mixtral.8*7&.55&.73&.57&.69&.45&.59&.46&.55\\
Mixtral 8*22&.69&.84&.72&\textbf{.83}&.64&.76&.66&.75\\
GPT-3.5&.66&.81&.69&.77&.54&.65&.56&.62\\
GPT-4&.\textbf{71}&\textbf{.85}&\textbf{.73}&\textbf{.83}&\textbf{.66}&\textbf{.79}&\textbf{.68}&\textbf{.77}\\
\bottomrule
\end{tabular}
\caption{Accuracy of the term extraction regardless of their assigned frame (left), and frame wise (right), for four different metrics: exact match (raw), one of the strings contained in the other (ovlp), lemmatized expression (lem) and lemmatized head noun (head). Sample size \(N=6050\). The highest accuracy scores appear highlighted in bold.}
\label{tab:results-extraction-no-position}
\end{table}

\begin{table}[t!]
\footnotesize
\centering
\begin{tabular}{l@{\hspace{2pt}}|@{\hspace{2pt}}l@{\hspace{2pt}}|@{\hspace{2pt}}c}
\toprule
\textbf{Example }&\multicolumn{2}{@{\hspace{0pt}}l}{
\textbf{Written laws} are like \textbf{spiders' webs}, }\\
&\multicolumn{2}{@{\hspace{0pt}}l}{ and will, like them, only entangle}\\
&\multicolumn{2}{@{\hspace{0pt}}l}{ and hold \textbf{the poor and weak}.}\\

\midrule
\textbf{Explicit}& T1: written laws,&\textbf{S2:}\\
\textbf{terms }& S1: spider's web&\\
& T2: the poor and weak & weak insects\\

\midrule
\textbf{Model}&\textbf{Generated S2 terms}&\textbf{Rating}\\
\midrule
Mixtral&spiders&0\\
8*24&entangle and hold&0\\
\midrule
Gpt-4&entangled beings;&1\\
&small insects&2\\
\bottomrule
\end{tabular}
\caption{Examples of generated arguments and human ratings for the implicit argument S2 with on example sentence }
\label{tab:examples-argument-generation}
\end{table}

\begin{table}[t!]
\footnotesize
\centering
\begin{tabular}{l@{\hspace{4pt}}l@{\hspace{5pt}}llll|ll}
\toprule
\textbf{In}&&\multicolumn{5}{c}{\textbf{Acc. for fields}}\\
\textbf{Input}&\textbf{Model}&\textbf{T1}&\textbf{T2}&\textbf{S1}&\textbf{S2}&\textbf{All}&\textbf{Q.}\\
\midrule
&Llama-3&-&.71&.81&.57&.71&.51\\
&Mixtral 8*7&-&.26&.50&.46&.41&.19\\
\textbf{T1}&Mixtral 8*22&-&.74&.79&.64&.73&.55\\
&GPT-3.5&-&.47&.45&.49&.47&.31\\
&GPT-4&-&.76&.89&.68&.78&.58\\
\midrule
 
&Llama-3&.84&-&.76&.60&.74&.59\\
&Mixtral 8*7&.75&-&.70&.55&.68&.51\\
\textbf{T2}& Mixtral 8*22&.84&-&.79&.64&.77&.64\\
&GPT-3.5&.85&-&.81&.61&.77&.59\\
&GPT-4&.88&-&.83&.67&.80&.64\\
\midrule

&Llama-3&.86&.70&-&.64&.74&.51\\
&Mixtral 8*7&.82&.40&-&.46&.58&.32\\
\textbf{S1}&Mixtral 8*22&.86&.74&-&.66&.76&.57\\
&GPT-3.5&.83&.65&-&.50&.67&.44\\
&GPT-4&.90&.75&-&.73&.81&.61\\
\midrule

&Llama-3&.68&.48&.78&-&.65&.47\\
&Mixtral 8*7&.78&.37&.52&-&.57&.32\\
\textbf{S2}& Mixtral8*22&.82&.67&.74&-&.75&.57\\
&GPT-3.5&.77&.50&.53&-&.60&.38\\
&GPT-4&.77&.63&.85&-&.76&.60\\
\midrule

&Llama-3&.79&.63&.78&.60&.71&.52\\
&Mixtral 8*7&.78&.34&.57&.49&.55&.34\\
\textbf{Agg.}&Mixtral8*22&.84&.72&.77&.65&.75&.58\\
&GPT-3.5&.81&.54&.60&.53&.62&.43\\
&GPT-4&.85&.71&.86&.69&.77&.61\\

\bottomrule
\end{tabular}
\caption{Accuracy results split by input and output frame. The accuracy for extraction of the four frames T1, T2, S1 and S2 individually and aggregated (All), and finally the accuracy results for extraction of the entire quadruple (Q). All scores are computed using the head-noun accuracy metric.}
\label{tab:results-all-slots}
\end{table}

\begin{table}[t!]
\footnotesize
\centering
\begin{tabular}{c@{\hspace{5pt}}|@{\hspace{5pt}}c@{\hspace{2pt}}c|@{\hspace{2pt}}c@{\hspace{2pt}}|c@{\hspace{2pt}}c@{\hspace{2pt}}|c@{\hspace{3pt}}c}
\toprule
\textbf{i}&\textbf{N}&\textbf{NQ}&\textbf{Llama-3}&\textbf{Mix. 8*7}&\textbf{8*22}&\textbf{GPT-3.5}&\textbf{4}\\

\midrule
0&3407&94&.70&.56&.75&.64&.79\\
1&2457&92&.73&.56&.77&.62&.78\\
2&186&10&.59&.56&.66&.58&.65\\
\bottomrule
\end{tabular}
\caption{Accuracy results split by the number of implicit terms per instance (i) using the head-nouns metric. N is the total number of runs and NQ the number of quadruples.}
\label{tab:results-implicit-terms}
\end{table}

\begin{figure}[t!]
\includegraphics[width=\columnwidth]{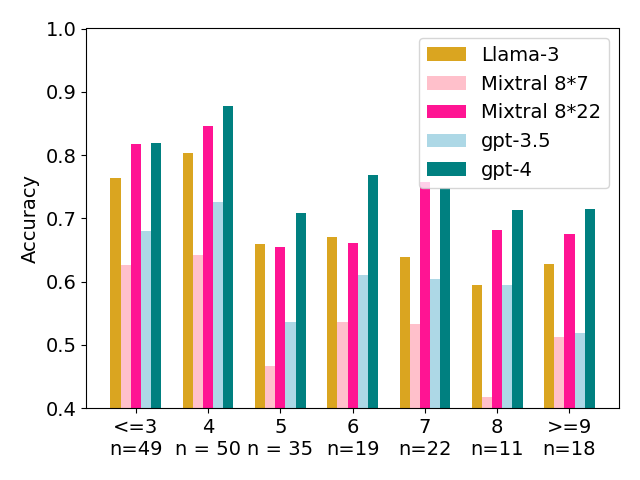}
\caption{Accuracy results of the Llama-3, gpt-3.5, gpt-4 and Mixtral models for various number of nouns in the sentence. n is the number of distinct sentence for each set}
\label{fig:accuracy-number-of-nouns}
\end{figure}

\begin{table}[t!]
\footnotesize
\centering
\begin{tabular}{l@{\hspace{5pt}}l@{\hspace{5pt}}l@{\hspace{5pt}}l@{\hspace{6pt}}l@{\hspace{6pt}}l@{\hspace{6pt}}l@{\hspace{6pt}}l}
\toprule
\textbf{Model}&\textbf{Tot.}&\textbf{Dedup.}&\textbf{Ave.}&\textbf{r=0}&\textbf{r=1}&\textbf{r=2}&\textbf{Score}\\

\midrule

Mixtral 8*22&356&153&6.8&191&25&140&0.75\\
GPT-4&369&188&7.1&132&21&216&1.21\\
\bottomrule
\end{tabular}
\caption{Results of the manual evaluation of the generated implicit terms of 50 sentences. The total number of evaluated terms (Tot.), number of distinct terms (Dedup.) and average number of evaluated term per implicit slot (Ave.) are shown in the first three columns. The relevance score appears in the last column. The rating \(r\) distribution is shown for the non deduplicated instances in the previous three columns.  }
\label{tab:results-argument-generation}
\end{table}

\section{Results}
\label{sec:results}
The models are evaluated on their performance in extracting analogies from literary texts, focusing on two distinct aspects: extracting the arguments of the analogy that can be found explicitly in the text, and generating a relevant concept when the argument is only implicitly suggested in the text\footnote{The data and scripts used in our experiments are available at \url{https://github.com/Mionies/metaphoric-analogies-extraction}.}.

\paragraph{Analogy extraction results for explicit terms}
The results with the four metrics are presented in Table \ref{tab:results-extraction-no-position}. On the left, we show the accuracy of each model per frame, ignoring whether the word has been extracted in the correct field (T1, T2, S1, S2), and on the right, we take this field into account. The lemmatized head noun-based matching function is retained for the remaining analysis because it allows flexibility in the quantity of modifiers extracted along the head nouns by the models and allows a variation between the plural and singular forms of the head noun, while restricting the answers counted as correct to the NP's that have identical core concepts as the human annotations.

Using the lemmatized head-noun match based metric, and considering each field T1, T2, S1 or S2 separately (in opposition to the accuracy computed on the entire quadruples Q), the mean accuracy obtained using GPT 3.5 over 10 batches of experiments with different input examples is 0.62 (standard deviation= 0.03) with a maximum value of 0.69 and minimum value of 0.58. Table \ref{tab:results-extraction-no-position} shows the accuracy averaged over 3 batches for all the models. GPT-4 reaches 77\% followed closely by the largest Mixtral model 8*22 with 75\%. Llama-3-70B and Mixtral 8*7, which have comparable sizes, obtain very different scores with 55\% for the smallest Mixtral and 16\% more for Llama-3. The gap between finding the correct concepts in the text regardless of its position in the mapping (frame-blind), and frame-wise narrows down when the models become better at the task. GPT-4 reaches an accuracy of 61\% for the correct extraction of all the explicit terms of a quadruple (see Table \ref{tab:results-all-slots}).

\paragraph{Are there significant differences in the performance between the frames?} Table \ref{tab:results-all-slots} shows the results split by provided input frame and generated output frame. The models perform better when they are given T2 as input. As for human evaluation, the models are also more successful at extracting T1 and S2 than T2 and S2, and S2 obtains the lowest accuracy. 

\paragraph{Does the number of implicit terms in the mapping or the number of nouns in the text impact the accuracy of the extraction performance?} Table \ref{tab:results-implicit-terms} shows very similar results for analogies that are fully explicit and analogies that have one implicit term. There are only 10 instances with two explicit terms in our test data. The size of this subset is too small to reach significance, but the performance decreases for these 10 examples.
Regarding the number of nouns in the text, Figure \ref{fig:accuracy-number-of-nouns} summarizes the results, with accuracy shown for sets of instances containing n<=3 to n>=9 nouns. Unsurprisingly, the best performance is obtained unsurprisingly on sentences containing exactly 4 nouns, often corresponding to the four elements to extract in fully explicitly formulated analogies. We observe that performance does not decline as the number of nouns increases, suggesting the potential for robustness of the extraction in the case of extending the task to longer texts.

\paragraph{Can the models generate relevant concepts when a term is implicit in the sentence?} While Mixtral 8*22 and GPT-4 are very close in accuracy for explicit term extraction, Table \ref{tab:results-argument-generation} shows a much larger gap for the quality of the generated terms. GPT-4 scores 1.21 with our relevance score, indicating that it generates relevant concepts more often than irrelevant ones, however, there is a lot of room for improvement.

\section{Discussion}
\label{sec:discussion}

\paragraph{Limitations of the evaluation method.} We rely on the lemmatized head-noun of the phrases for matching the concepts extracted by the models to the human answer in the evaluation. This is an imperfect metric that does not take into account all possible reformulations that could be produced by the model. For example, given the sentence \textit{All the world's a stage, and all the men and women merely players.} and the concept $T1=World$, an answer from GPT-4 is ($T2:People, S1: Stage, S2:Actors$). This answer could be considered correct because the meaning of the analogy is preserved, but GPT-4 reformulates two of the three elements: \textit{players} become \textit{actors} and \textit{the men and women} become \textit{people}. The results reported in our experiments are therefore to be considered as the lowest boundary of the performance of the models for analogy extraction\footnote{ The instruction given in the prompt (see Section \ref{sec:analogy-extraction}), is to generate a new concept only when the relevant concept does not appear explicitly in the text.}.

\paragraph{Possible Extensions} 
There are several straightforward possible directions for enriching the approach and improving performance. Increasing the number of prompt examples may of course lead to a better accuracy. A step-by-step approach where the extraction of the four terms of the analogy would be separated from the task of deciding which frame T1, T2, S1 or S2 belongs to might fill the gap between the frame-blind and frame-aware performances shown in Table \ref{tab:results-extraction-no-position}. The effect of model hyper-parameters such as temperature could also be systematically tested.

\section{Conclusion}

In this paper, we introduce a new task and dataset centered on extracting metaphoric analogies from free text. The results showed that language models can become valuable tools for the conversion of unstructured metaphors to structured analogical concept mapping. The performance of models analysed (in particular GPT-4 and Mixtral) is remarkably high, in line with human annotators. We believe that the results can even be further improved with current models by, for example, performing prompt engineering or fine-tuning, which we did not attempt in this paper. These results may open up several avenues for future research on metaphor extraction and understanding. 

In future work we would like to integrate the relation between concepts in our framework and extend it to different structure mappings. We are also interested in testing the utility and validity of the extracted 4-term mappings by LLMs in open text, which is likely to be more complicated as it would at the minimum entail metaphor identification.

\section*{Limitations}
Our experiments are run on a novel dataset that has a relatively small size (204 instances). The dataset instances and the implicit terms generated by the GPT-4 and Mixtral 8*22 models that are manually evaluated in Section \ref{sec:results} receive in majority one single annotation. The reason for this small scale experiment lies in a limited budget and the difficulty of recruiting expert annotators that are qualified to complete this complex task. We chose qualitative annotation from experts instead of a large number of annotations after three non-expert students initially asked to complete the task declared themselves unable to solve it, which led us to rely mainly on an annotator aligned with the majority vote for the analogy frame extraction task. The inter-annotator agreement computed for the sentence tagging and implicit terms rating tasks gives us confidence in the meaningfulness of the task in spite of this limitation.

Another limitation of our study is the absence of manual evaluation of analogy frame extraction. We consider the metrics used to evaluate accuracy scores to be a lower bound of the true accuracy of the model. It would be interesting to complete the analysis with a manual rating of the extracted terms, enabling a comparison between the true accuracy and the head-noun based metric scores, and a categorization of the errors made by the model.

\section*{Ethical Statement}

We have not identified any important ethical issues within this work. As usual with LLMs, there is the risk that the generated output may be misleading or incorrect, so interpretation of the results and any potential deployment should be carried out with caution. The dataset created contains sentences that may be viewed as offensive or harmful.

\bibliography{anthology-shrunk,custom}

\section*{Acknowledgments}
We thank the anonymous reviewers for their invaluable feedback, all the annotators who participated in the project, Thomas Green at the Advanced Research Computing at Cardiff (ARCCA) for giving us access to the computational resources. Jose Camacho-Collados and Dimosthenis Antypas are supported by a UKRI Future Leaders Fellowship.

\appendix

\section{Dataset annotations and description additional information}
\label{appendix:datasets}
Complementary information about the dataset characteristics and annotation metrics are provided here.

\subsection{Inter-annotator agreement table }
\label{appendix:iaa-without-implicit}

In Table \ref{tab:appendix-iaa-without-implicit}, agreement is computed after removing all the fields that a majority of annotators labelled as implicit from the evaluation.

\begin{table}[]
\footnotesize
\begin{tabular}{ll@{\hspace{9pt}}l@{\hspace{9pt}}l@{\hspace{9pt}}l@{\hspace{24pt}}l@{\hspace{9pt}}l@{\hspace{9pt}}l@{\hspace{9pt}}l}
\toprule

\textbf{Anno.}&\multicolumn{4}{c}{\textbf{Exact match}}&\multicolumn{4}{c}{\textbf{Head noun match}}\\

&\textbf{T2}&\textbf{S1}&\textbf{S2}&\textbf{all}&\textbf{T2}&\textbf{S1}&\textbf{S2}&\textbf{all}\\
\midrule

\textbf{A1}& .55&	.78&	.48& .52& .55& .78& .49& .53\\
\textbf{A2}& .58&	.84&	.53& .53& .58& .84& .60& .54\\
\textbf{A3}& .70&	.88&	.48& .57& .70& .88& .50& .57\\
\textbf{A4}& .75&	.91&	.62& .64& .75& .91& .68& .65\\
\textbf{A5}& .77&	.91&	.60& .66& .77& .91& .64& .67\\
\midrule
\textbf{av.}& .67& .86& .54& .58& .67& .86& .58& .59\\
\bottomrule
\end{tabular}
\caption{Averaged pairwise inter-annotator agreement scores for the extraction of the explicit terms, with Cohen's Kappa. The analogy fields (T2, S1 or S2) values that are considered implicit by a majority of annotators are discarded from the calculation. The agreement is shown by frame and aggregated for all the frames.}
\label{tab:appendix-iaa-without-implicit}

\end{table}

\subsection{Histograms of the distribution of the number of nouns on the dataset.}
\label{appendix-histograms}

\begin{figure}[]
\includegraphics[width=\columnwidth]{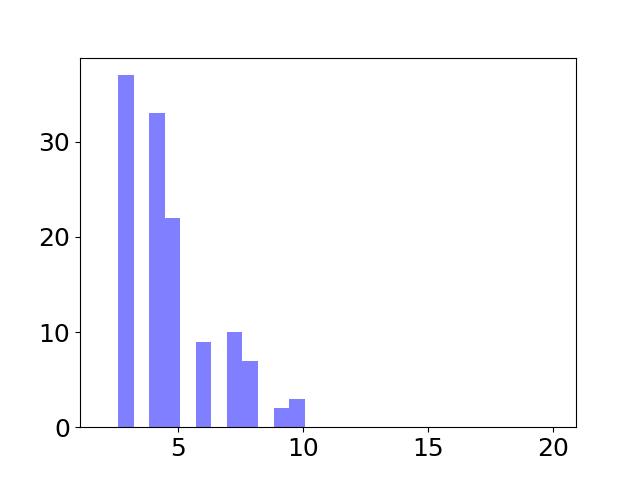}
\caption{Histogram of the number of nouns per instance in the entire dataset and for the  main source of the dataset Grothe }
\label{fig:histogram-grothe}
\end{figure}

\begin{figure}[]
\includegraphics[width=\columnwidth]{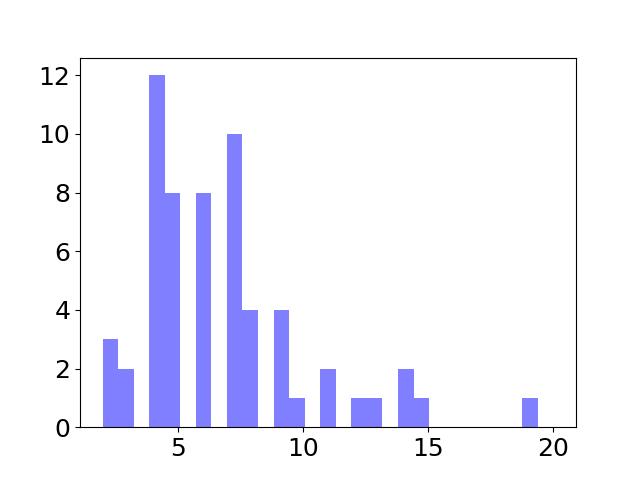}
\caption{Histogram of the number of nouns per instance in the entire dataset and for Pasanek texts}
\label{fig:histogram-pasanek}
\end{figure}

Figure \ref{fig:histogram-grothe} and \ref{fig:histogram-pasanek} are the histograms showing for distribution of the data from each source ~\cite{grothe} and ~\cite{metaphorsofmind:2015} of the number of nouns in each instance.

\subsection{Distribution of the implicit terms per frame and per source }

\begin{table}[t!]
\footnotesize
\begin{tabular}{l@{\hspace{8pt}}l@{\hspace{8pt}}l@{\hspace{16pt}}l@{\hspace{8pt}}l@{\hspace{16pt}}l@{\hspace{8pt}}l@{\hspace{16pt}}l@{\hspace{8pt}}l}
\toprule
\textbf{Source}&
%\textbf{}&
\multicolumn{2}{l}{\textbf{T1}}&
\multicolumn{2}{l}{\textbf{T2}}&
\multicolumn{2}{l}{\textbf{S1}}&
\multicolumn{2}{l}{\textbf{S2}}\\

&
%\textbf{\(i>0\)}&
\textbf{\(n\)}&\textbf{\(\%\)}&
\textbf{\(n\)}&\textbf{\(\%\)}&
\textbf{\(n\)}&\textbf{\(\%\)}&
\textbf{\(n\)}&\textbf{\(\%\)}\\
\midrule

Grothe&3&2.4&21&16.9&3&2.4&40&32.3\\
Pasanek& 2&3.3& 12&20& 11 &18.3&16&26.7\\
Other&2&10&6&30&0&0&0&0\\
\midrule
\textbf{Total}&7&3.4&39& 19.1 &14&6.9&56&27.5\\
\bottomrule
\end{tabular}
\caption{Number and percentage of implicit terms among all instances for each frame of the analogies T1, T2, S1 and S2.}
\label{tab:implicit-terms}
\end{table}

\section{Prompts}
\label{sec:appendix-llama3-prompt}

\subsection{Prompt text}
\label{appendix-prompts}
\begin{tcolorbox}
\footnotesize

Let T1, T2, S1 and S2 be four concepts forming a metaphor in a short text. The relation between the concepts T1 and T2 is analogous to the relation between the concepts 
S1 and S2. The two concepts T1 and T2 belong to the target domain of the metaphor, they express the main topic discussed in the text. The two concepts S1 and S2 belong to the source domain of the metaphor, they express the image of the metaphor. Given a short text that contains a metaphor, and one of the four concepts, your task is to find three other concepts forming a metaphor with it in the text. The provided concept and the three extracted concepts must together form an analogy. Sometimes, T1, T2, S1 
or S2 might be implicit in the text (the word might not appear in the text),and in this case you should infer a correct concept.                                         
\bigbreak

\noindent\textbf{Example 1}:
\bigbreak
\noindent\textbf{Text containing a metaphor:} "Zeal is a volcano, on the peak of which the grass of indecisiveness does not grow."

\noindent\textbf{Concept T1}: "zeal"     
\noindent\textbf{Answer:}                       

\noindent\textbf{T1}: zeal  

\noindent\textbf{T2}: indecisiveness 

\noindent\textbf{S1}: volcano 

\noindent\textbf{S2}: grass                                                                                             
\bigbreak

\noindent\textbf{\(<\) 6 additional examples are provided here \(>\)}
\bigbreak
Now it is your turn. Find four concepts forming a metaphor in the following sentence, given T1.
\bigbreak
\noindent\textbf{\(<\) 1 example and 1 concept are provided here \(>\)}

\end{tcolorbox}

\subsection{ System prompt for the experiments done with Llama-3}

Minor modifications of the prompt used for the OpenAI GPT-3.5 and GPT-4 experiments are done for the Llama-3-instruct model experiments to better control the output format of the generated text. Here is the exact prompt used for the Llama-3- instruct model experiments. A system prompt is added ahead of the user prompt : 

\begin{tcolorbox}
\footnotesize
 You always answer by providing the four expressions named T1, T2, S1 and S2, outputting one expression per line.
 \end{tcolorbox}

\end{document}